\documentclass[sigconf]{acmart}

\usepackage{amsmath,amsthm}
\usepackage{microtype}
\usepackage{algorithm}
\usepackage{algpseudocode}
\usepackage{graphicx}

\AtBeginDocument{%
  }

\setcopyright{acmlicensed}
\copyrightyear{2018}
\acmYear{2018}
\acmDOI{XXXXXXX.XXXXXXX}
\acmConference[Conference acronym 'XX]{Make sure to enter the correct
  conference title from your rights confirmation email}{June 03--05,
  2018}{Woodstock, NY}
\acmISBN{978-1-4503-XXXX-X/2018/06}

\begin{document}
\emergencystretch=1.5em

\title{SAGG: Sample-Adaptive Gradient Gating for Robust Multimodal Learning under Heterogeneous Corruption}

\author{Wentao Zhang}
\affiliation{%
  \institution{Tsinghua University}
  \city{Shenzhen}
  \country{China}
}
\email{zhang-wt24@mails.tsinghua.edu.cn}

\author{Yifan Zhu}
\authornote{Corresponding author.}
\affiliation{%
  \institution{Sichuan University}
  \city{Chengdu}
  \country{China}
}
\email{zyfan@stu.scu.edu.cn}

\author{Yutong Zhang}
\affiliation{%
  \institution{Sichuan University}
  \city{Chengdu}
  \country{China}
}
\email{yutongzhang@stu.scu.edu.cn}

\author{Wentao Mo}
\affiliation{%
  \institution{Tsinghua University}
  \city{Shenzhen}
  \country{China}
}
\email{mow10@mails.tsinghua.edu.cn}

\renewcommand{\shortauthors}{Zhang et al.}

\begin{abstract}
Multimodal gradient balancing methods modulate encoder gradients with a shared scalar per modality, implicitly assuming that corruption is uniform across the training batch. In practice, corruption is sample-heterogeneous: within a single mini-batch, different samples may have different modalities corrupted. We prove that under this heterogeneous corruption model, any batch-level sample-agnostic linear estimator with a shared modulation parameter incurs an irreducible bias with respect to the clean-data gradient, and that sample-level all-or-nothing gating is the unique unbiased strategy within a natural distribution-free estimator class. Motivated by this result, we propose Sample-Adaptive Gradient Gating (SAGG), which makes a binary retain-or-discard decision per sample via an online feature-norm quality test and incorporates a truncation mechanism for variance control. We prove that SAGG-based SGD converges at the standard $\mathcal{O}(1/\sqrt{T})$ rate to stationary points of the clean loss without a corruption-dependent error floor, and derive a certified robustness radius for the independent-encoder architecture that connects per-modality Lipschitz constants to the classification margin. Experiments on Kinetics-Sounds and UCF-101 under Gaussian noise injection, partial modality missing, and natural contribution imbalance show that SAGG consistently outperforms ten existing methods, with the largest gains in high-corruption regimes where batch-level bias is most~severe.
\end{abstract}



\maketitle

\section{Introduction}
Multimodal learning combines complementary signals from heterogeneous sources to construct representations richer than any single modality can provide \cite{baltruvsaitis2018multimodal,chen2024next}. However, real-world deployments routinely encounter heterogeneous corruption: sensor-specific noise \cite{mao2023robust,radenovic2023filtering}, missing modalities \cite{wudeep}, and intrinsic quality disparities \cite{zhang2024multimodal}. Critically, these corruptions do not distribute uniformly; within a single training batch, one sample may have clean visual input but corrupted audio, while the next exhibits the opposite pattern.

A prominent line of research addresses modality imbalance through gradient-level intervention. OGM \cite{peng2022balanced}, GMML \cite{zhang2025gmml}, CGGM \cite{guo2024classifier}, and Sun et al. \cite{sun2021learning} all compute a single modulation weight per modality and apply it uniformly to every sample in the batch. However, there is a fundamental mismatch between such batch-level control and sample-level corruption. When a fraction of samples have corrupted audio while the rest are clean, a batch-level estimator must choose one scaling factor $\kappa$ for the audio encoder gradient, blending clean and corrupted signals indiscriminately. No single $\kappa$ can eliminate the resulting bias across all parameter configurations simultaneously (Section~\ref{sec:theory}), indicating that the error is structural rather than a tuning artifact.

The natural resolution is to match control granularity to corruption granularity. Because the classification loss couples all modalities through the fused representation, corruption in any one modality contaminates the entire per-sample gradient. Therefore, the natural unit of gradient control is the sample: either all modalities are intact and the gradient is trustworthy, or at least one is compromised and the gradient should be excluded.

We propose Sample-Adaptive Gradient Gating (SAGG), a framework that makes a binary retain-or-discard decision for each sample based on an online quality assessment, together with a truncation mechanism for variance control. We show that, within a natural class of distribution-free linear estimators, unbiased estimation uniquely requires sample-level all-or-nothing gating. SAGG-based SGD converges at rate $O(1/\sqrt{T})$ to a stationary point of the clean loss $L_{\mathrm{cln}}$ with no constant-order residual bias; the only cost is a variance factor $(1{-}\rho)^{-1}$ from the reduced effective batch size. We further derive a certified robustness condition for multimodal models expressed through per-modality Lipschitz constants and the classification margin. Experiments on Kinetics-Sounds and UCF-101 under noise injection, partial missing, and natural imbalance confirm that SAGG consistently outperforms prior methods \cite{peng2022balanced,wei2024fly,wei2024diagnosing,zhang2025gmml,reconboost}.

Our contributions are as follows:
\begin{itemize}
    \item We identify a granularity mismatch in existing batch-level gradient modulation under heterogeneous multimodal corruption, and show theoretically that this mismatch precludes unbiased estimation within a natural estimator class unless sample-level all-or-nothing gating is used.
    \item We propose SAGG, a sample-adaptive gradient gating framework with a truncation mechanism for variance control, and prove that SAGG-based SGD converges at rate $O(1/\sqrt{T})$ to a stationary point of the clean loss without constant-order corruption-induced residuals.
    \item We provide a certified robustness framework for the independent-encoder architecture as an analytical tool to evaluate the robustness benefits of corruption-free training, and validate this connection experimentally.
    \item Extensive experiments on KS and UCF-101 demonstrate that SAGG achieves state-of-the-art performance across noise, missing modality, and contribution imbalance scenarios, with particular advantages under heterogeneous within-batch corruption.
\end{itemize}

\section{Related Work}

\subsection{Multimodal Fusion and Robustness under Corruption}

Multimodal learning fuses modality-specific encoder outputs before a shared classifier \cite{baltruvsaitis2018multimodal,chen2024next}, with strategies ranging from early concatenation \cite{ngiam2011multimodal} to late combination \cite{atrey2010multimodal} and intermediate cross-modal interaction \cite{zadeh2017tensor}. Real-world corruption includes sensor noise \cite{mao2023robust,radenovic2023filtering} addressed by attention weighting \cite{tsai2019multimodal} and confidence-aware fusion \cite{han2022trusted}; missing modalities \cite{wudeep} handled via imputation \cite{tran2017missing,ma2021smil} and prompt adaptation \cite{lee2023multimodal}; and contribution imbalance \cite{peng2022balanced,fan2023pmr} mitigated through gradient regulation \cite{sun2021learning} and Pareto formulations \cite{mmpareto}. These approaches generally assume corruption patterns uniform within a batch. In contrast, our work targets sample-heterogeneous corruption where the pattern varies from sample to sample within a mini-batch. Robustness interventions have also been explored at inference time in embodied multimodal systems: VLA-SCT \cite{zhang2026knowing} employs a training-free self-correction loop for action refinement and termination detection in vision-language-action models. This is complementary to SAGG, which addresses corruption-induced gradient bias during multimodal training.

\subsection{Gradient-Level Modality Balancing}

OGM \cite{peng2022balanced} pioneered gradient-level balancing by attenuating the dominant modality's gradient based on classifier confidences. Extensions include OPM \cite{wei2024fly} (phase-adaptive modulation), DRBM \cite{wei2024diagnosing} (imbalance diagnosis), CGGM \cite{guo2024classifier} (parameter-level correction), PMR \cite{fan2023pmr} (prototype-based rebalancing), Reconboost \cite{reconboost} (boosting formulation), MMPareto \cite{mmpareto} (Pareto-optimal solutions), and GMML \cite{zhang2025gmml} (symmetric hyperbolic-tangent weighting with $\ell_2$-norm constraints). All these methods apply a shared modulation weight uniformly to every sample in the mini-batch. Under heterogeneous corruption where modality reliability varies across samples, this creates a granularity mismatch leading to irreducible bias (Section~\ref{sec:bias}). SAGG departs from this paradigm by making a binary retain-or-discard decision per sample.

\subsection{Sample-Level Reweighting and Robust Optimization}

Sample-level reweighting assigns individual weights reflecting each sample's quality. Curriculum learning \cite{bengio2009curriculum} and self-paced learning \cite{kumar2010self} order samples by difficulty; meta-learning reweighting \cite{shu2019meta} optimizes on a clean validation set; and noise-robust methods such as Co-teaching \cite{han2018coteaching} and loss-based selection \cite{arazo2019unsupervised} down-weight mislabeled samples. DRO \cite{duchi2021learning} and Group DRO \cite{sagawa2020distributionally} optimize over worst-case distributions. These methods share with SAGG the principle of treating samples differently, but most assign continuous weights, whereas SAGG employs a hard binary gate: cross-modal gradient coupling means corruption in any modality contaminates the entire per-sample gradient, so partial down-weighting cannot fully eliminate bias. SAGG combines sample-level granularity with a corruption-aware gating mechanism, supported by formal guarantees on unbiasedness and convergence.

\begin{figure*}[t]
\centering
\includegraphics[width=0.80\textwidth]{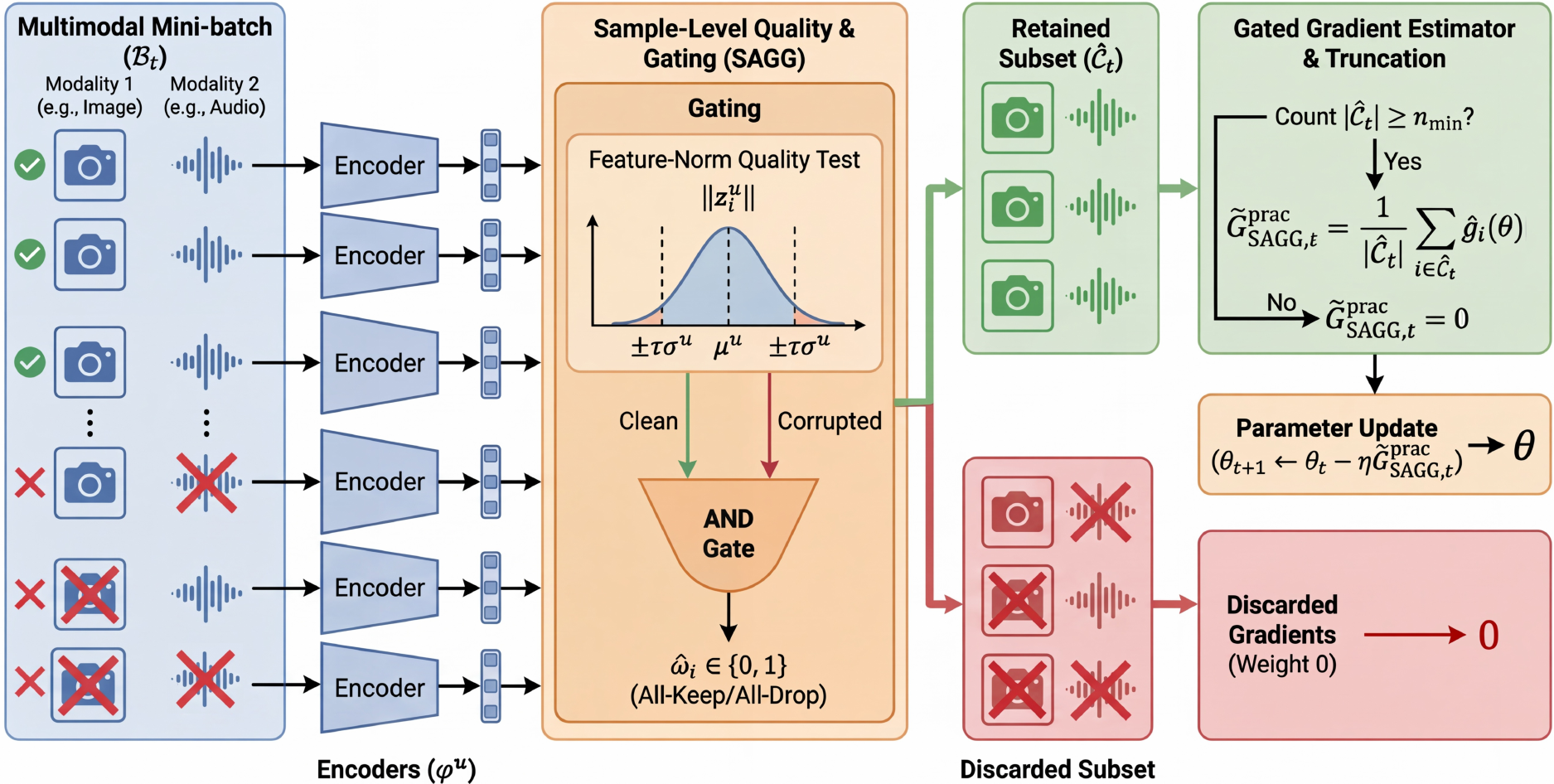}
\caption{Overview of the SAGG framework. For each sample in a mini-batch, modality-specific encoders produce feature representations whose norms are compared against EMA-tracked statistics. A sample-level quality gate makes a binary retain-or-discard decision: only samples with all modalities passing the feature-norm test contribute to the gradient update, while samples with any corrupted modality are entirely excluded. A truncation mechanism ensures variance control when the retained subset is small.}
\label{fig:architecture}
\end{figure*}

\section{Preliminaries}
\subsection{Multi-modal Learning}
\label{3.1}
We adopt a standard joint multimodal architecture with modality-specific encoders and a shared classifier. Consider a dataset $\mathcal{D} = \{(x_i, y_i)\}_{i=1}^N$ with $M$ modalities ($M \geq 2$), where $x_i = (x_i^1, \ldots, x_i^M)$ is the input sample and $y_i \in [K]$ is the label. We employ $M$ independent encoders $\varphi^u(\theta^u, \cdot)$ ($u \in [M]$) and a shared linear classifier with weight $W$ and bias $b$. The model output is:
\begin{equation}\label{encoder}
f(\theta, x_i) = W[\varphi^1(\theta^1, x_i^1); \ldots; \varphi^M(\theta^M, x_i^M)] + b \end{equation}
where $\theta = (\theta^1, \ldots, \theta^M, W, b)$ collects all learnable parameters. Let $z_i^u = \varphi^u(\theta^u, x_i^u)$ denote the encoded feature of modality $u$. By the chain rule, the gradient of the per-sample loss with respect to encoder $u$ is:
\begin{equation}\label{gradient}\nabla_{\theta^u}\ell_i = \left(\frac{\partial z_i^u}{\partial \theta^u}\right)^\top \nabla_{z_i^u} \ell_i \end{equation}
where $\nabla_{z_i^u} \ell_i$ is determined by the fused representation of all modalities through the shared classifier and loss. Consequently, corruption in any single modality may affect the gradients of all encoders. This cross-modal coupling suggests that robust gradient estimation may not be achieved by treating each modality independently, and motivates estimators that account for sample-level corruption heterogeneity.

\subsection{Heterogeneous Corruption Model}
\label{3.2}
Real-world multimodal systems face heterogeneous corruption where different samples exhibit different corruption patterns simultaneously. We formalize this as follows.
\begin{definition}[Heterogeneous Corruption] For each sample $i$ and modality $u$, define the clean set $\mathcal{C}^u \subseteq [N]$ and corrupted set $\mathcal{N}^u = [N] \setminus \mathcal{C}^u$ as a latent partition used for theoretical analysis; these sets are not assumed to be observable during training. Cleanliness is defined with respect to whether the observed training input of modality $u$ coincides with the true input. The observed input is:
\begin{equation}
\hat{x}_i^u = \begin{cases} x_i^u & i \in \mathcal{C}^u \\ c_i^u(x_i^u) & i \in \mathcal{N}^u \end{cases}\end{equation}
\end{definition}
where $c_i^u: \mathcal{X}^u \to \mathcal{X}^u$ is an arbitrary sample-dependent corruption operator (additive noise: $c_i^u(x_i^u) = x_i^u + \epsilon_i^u$; modality missing: $c_i^u(x_i^u) = 0$). The global clean set is $\mathcal{C} = \bigcap_{u=1}^M \mathcal{C}^u$ and the corruption ratio is $\rho = 1 - |\mathcal{C}|/N$. For a mini-batch $\mathcal{B}_t$ of size $B$, let $\mathcal{C}_t := \mathcal{B}_t \cap \mathcal{C}$ denote the globally clean subset. The reference objective is the empirical clean loss:
\begin{equation}
L_{\mathrm{cln}}(\theta) = \frac{1}{|\mathcal{C}|}\sum_{i \in \mathcal{C}} \ell(f(\theta, x_i), y_i) \end{equation}

\subsection{Batch-Level Gradient Modulation}
\label{3.3}
Existing multimodal gradient balancing methods compute modality- or batch-level statistics and use them to rescale gradients with coefficients shared across the mini-batch. We abstract this shared-rescaling structure into the following general form.

\begin{definition}[Batch-Level Sample-Agnostic Estimator]\label{d2} Denote the observed gradient of sample $i$ by $\hat{g}_i(\theta) := \nabla_\theta \ell(f(\theta, \hat{x}_i), y_i)$. A gradient estimator is batch-level sample-agnostic if it takes the form:
\begin{equation}
\hat{G}_{\mathrm{batch}} = \frac{1}{B}\sum_{i \in \mathcal{B}_t} \kappa_t \odot \hat{g}_i(\theta) \end{equation}
where $\kappa_t \in \mathbb{R}^d$ ($d = \dim(\theta)$) is a modulation vector shared across all samples in $\mathcal{B}_t$, depending on batch observations but not on sample-specific corruption indicators. This captures the shared-rescaling structure of OGM, GMML, and related methods. Under heterogeneous corruption, such estimators can be biased with respect to $L_{\mathrm{cln}}$ (Theorem 1).
\end{definition}

\section{SAGG: Proposed Method}

We propose Sample-Adaptive Gradient Gating (SAGG), which departs from batch-level shared rescaling and instead constructs a sample-level retained subset over which gradients are averaged. We introduce an oracle gating estimator (Section \ref{4.1}), a truncated variant for variance control (Section \ref{4.2}), and a practical approximation based on estimated sample quality (Section \ref{4.3}). The overall framework is illustrated in Figure~\ref{fig:architecture}.
\subsection{Sample-Adaptive Gradient Gating}
\label{4.1}
We begin by defining the clean per-sample gradient. For any sample $i \in [N]$, let
\begin{equation}
g_i(\theta) := \nabla_\theta \ell(f(\theta, x_i), y_i) \end{equation}
denote the gradient computed from the true (uncorrupted) input $x_i$. For $i \in \mathcal{C}_t$, all modalities are clean, so $\hat{x}_i = x_i$ and hence $g_i(\theta) = \hat{g}_i(\theta)$. SAGG applies an all-keep/all-drop binary inclusion rule at the sample level: each sample in the mini-batch is either fully retained (all modalities clean) or fully discarded (at least one modality corrupted). Formally, define the gating indicator $\omega_i = \mathbf{1}[i \in \mathcal{C}_t]$. The SAGG gradient estimator is:
\begin{equation}\label{SAGG}\hat{G}_{\mathrm{SAGG}} = \begin{cases} \frac{1}{|\mathcal{C}_t|}\sum_{i \in \mathcal{C}_t} g_i(\theta) & |\mathcal{C}_t| \geq 1 \\ 0 & |\mathcal{C}_t| = 0 \end{cases} \end{equation}
That is, the estimator averages clean gradients exclusively over the globally clean subset $\mathcal{C}_t$, assigning zero weight to any sample with at least one corrupted modality.

Unlike batch-level methods that continuously rescale gradients via shared modality statistics, SAGG employs a hard binary gate. This design is driven by the cross-modal gradient coupling identified in Eq.~\ref{gradient}: since $\nabla_{z_i^u}\ell_i$ depends on all modalities through the fused prediction, partially corrupted samples can induce biased gradients for every encoder, not just the corrupted modality's encoder. The all-keep/all-drop rule ensures that only fully reliable gradient signals participate in the update.

A natural alternative is modality-level gating: for each modality $u$, retain sample $i$ only if modality $u$ is clean, allowing different subsets per modality. While this preserves more data, it does not guarantee alignment with $\nabla L_{\mathrm{cln}}$ due to the cross-modal coupling in $\nabla_{z_i^u}\ell_i$. Sample-level all-keep/all-drop gating is the estimator structure supported by our unbiasedness analysis in Theorem~2.
\subsection{Truncated SAGG Estimator}
\label{4.2}
When the retained clean subset is very small, the SAGG estimator (Eq. \ref{SAGG}) may exhibit high variance. To control this, we introduce a truncation mechanism. We use $n_{\min}$ to denote the truncation threshold. In the analysis, we instantiate it as $n_{\min} = \lceil B(1-\rho)/2 \rceil$; in practice, it is treated as a tunable hyperparameter. The truncated SAGG estimator is:
\begin{equation}
\label{T-SAGG}
\tilde{G}_{\mathrm{SAGG},t} = \begin{cases} \frac{1}{|\mathcal{C}_t|}\sum_{i \in \mathcal{C}_t} g_i(\theta) & |\mathcal{C}_t| \geq n_{\min} \\ 0 & |\mathcal{C}_t| < n_{\min} \end{cases} \end{equation}
The parameter update rule is:
\begin{equation}\theta_{t+1} = \theta_t - \eta \tilde{G}_{\mathrm{SAGG},t} \end{equation}
The probability of truncation and the bias it introduces are analyzed formally in Theorem 3.

By excluding corrupted samples, SAGG reduces the effective batch size from $B$ to approximately $B(1-\rho)$ in expectation. This typically increases gradient variance relative to standard SGD on clean data, reflecting the cost of discarding corrupted samples. The precise tradeoff between bias elimination and variance increase is analyzed in Theorem 3. We emphasize that Eqs. \ref{SAGG}–\ref{T-SAGG} define the oracle SAGG estimator, which assumes access to the true clean partition $\mathcal{C}_t$. The practical counterpart, which replaces the oracle with an estimated partition, is introduced next.

\subsection{Practical SAGG with Quality Estimation}
\label{4.3}
The oracle SAGG estimator requires identifying $\mathcal{C}_t$, which is not directly observable during training. We approximate it via a quality estimator $\hat{q}: \mathcal{X}^1 \times \cdots \times \mathcal{X}^M \to \{0, 1\}$ that predicts whether all modalities of a sample are clean:
\begin{equation}
\hat{\omega}_i = \hat{q}(\hat{x}_i^1, \ldots, \hat{x}_i^M) \end{equation}
The estimated clean subset is $\hat{\mathcal{C}}_t = \{i \in \mathcal{B}_t : \hat{\omega}_i = 1\}$. The practical SAGG estimator replaces both the oracle partition and the gradient object:
\begin{equation}
\tilde{G}_{\mathrm{SAGG},t}^{\mathrm{prac}} = \begin{cases} \frac{1}{|\hat{\mathcal{C}}_t|}\sum_{i \in \hat{\mathcal{C}}_t} \hat{g}_i(\theta) & |\hat{\mathcal{C}}_t| \geq n_{\min} \\ 0 & |\hat{\mathcal{C}}_t| < n_{\min} \end{cases}\end{equation}

Note that $\tilde{G}_{\mathrm{SAGG},t}^{\mathrm{prac}}$ uses observed gradients
$$\hat{g}_i(\theta) = \nabla_\theta \ell(f(\theta, \hat{x}_i), y_i)$$
over the estimated clean set~$\hat{\mathcal{C}}_t$, whereas the oracle version (Eq.~\ref{T-SAGG}) uses true clean gradients $g_i(\theta)$ over~$\mathcal{C}_t$. Unless otherwise stated, the convergence and unbiasedness results in Section~\ref{sec:theory} are established for the oracle~estimator. 

We instantiate the gate using a feature-norm heuristic that is architecture-agnostic and requires no auxiliary reconstruction modules or confidence heads. For each modality $u$, the encoder feature norm $\|z_i^u\| = \|\varphi^u(\theta^u, \hat{x}_i^u)\|$ serves as a corruption indicator: corrupted or missing inputs often produce abnormal feature norms.

We maintain exponential moving average (EMA) statistics of feature norms: denoting the batch mean as $\bar{n}_t^u = B^{-1}\sum_{i \in \mathcal{B}_t}\|z_i^u\|$, the EMA statistics are updated as $\mu^u \leftarrow \gamma\mu^u + (1{-}\gamma)\bar{n}_t^u$ (analogously for~$\sigma^u$), where $\gamma \in (0,1)$ is the EMA~momentum. We flag modality $u$ of sample $i$ as corrupted if $\|z_i^u\|$ falls outside $[\mu^u - \tau\sigma^u,\, \mu^u + \tau\sigma^u]$, and implement the sample-level all-keep/all-drop rule via a logical AND across modality-wise~checks:
\begin{equation}
\hat{\omega}_i = \prod_{u=1}^{M} \mathbf{1}\left[\|z_i^u\| \in [\mu^u - \tau\sigma^u,\, \mu^u + \tau\sigma^u]\right] \end{equation}
where time indices on $\mu^u, \sigma^u$ are omitted for simplicity. Other quality estimation strategies (e.g., reconstruction-based or confidence-based detectors) can be substituted without changing the SAGG framework.

When the quality estimator is imperfect, false positives, which admit corrupted samples as clean, reintroduce bias by including corrupted gradients, whereas false negatives, which reject clean samples, further reduce the retained sample count. A formal analysis of this degradation under estimation errors is left to future work. The complete training procedure is summarized in Algorithm \ref{alg:sagg_practical}.

\begin{algorithm}[t]
\caption{Practical training procedure of SAGG}
\label{alg:sagg_practical}
\begin{algorithmic}[1]
\Require Training dataset $\mathcal{D}=\{(x_i,y_i)\}_{i=1}^N$; total iterations $T$; learning rate $\eta$; threshold $\tau$; batch size $B$; truncation threshold $n_{\min}$; EMA momentum $\gamma$
\Ensure Optimized model parameters $\theta$
\State Initialize EMA statistics $\mu^u \gets 0$, $\sigma^u \gets 1$ for each modality $u\in[M]$
\For{$t=0$ to $T-1$}
    \State Sample mini-batch $\mathcal{B}_t$ from $\mathcal{D}$
    \State Compute $z_i^u=\varphi^u(\theta^u,\hat{x}_i^u)$ for all $i\in\mathcal{B}_t$, $u\in[M]$
    \State Compute $\bar{n}_t^u=\frac{1}{B}\sum_{i\in\mathcal{B}_t}\|z_i^u\|$ for each $u$
    \State Update $\mu^u \gets \gamma\mu^u+(1-\gamma)\bar{n}_t^u$, and update $\sigma^u$ analogously via batch second moment
    \State Estimate clean subset $\hat{\mathcal{C}}_t$ via the feature-norm gate
    \If{$|\hat{\mathcal{C}}_t| \ge n_{\min}$}
        \State $\tilde{G}_{\mathrm{SAGG},t}^{\mathrm{prac}} \gets \frac{1}{|\hat{\mathcal{C}}_t|}\sum_{i\in\hat{\mathcal{C}}_t}\hat{g}_i(\theta)$
    \Else
        \State $\tilde{G}_{\mathrm{SAGG},t}^{\mathrm{prac}} \gets 0$
    \EndIf
    \State $\theta \gets \theta - \eta \tilde{G}_{\mathrm{SAGG},t}^{\mathrm{prac}}$
\EndFor
\end{algorithmic}
\end{algorithm}

\section{Theoretical Analysis}
\label{sec:theory}
Our analysis establishes irreducible batch-level bias and oracle SAGG's uniqueness within a linear-estimator class (Sections~\ref{sec:bias}--\ref{sec:uniqueness}); proves residual-free $\mathcal{O}(1/\sqrt{T})$ convergence for the truncated estimator (Section~\ref{sec:convergence}); and derives a certified radius (Section~\ref{sec:robustness}). Imperfect gates remain future work.

\textbf{Assumption 1.} $L_{\mathrm{cln}}(\theta)$ is $\beta$-smooth: for all $\theta,\theta'$,
\begin{equation*}
\|\nabla L_{\mathrm{cln}}(\theta) - \nabla L_{\mathrm{cln}}(\theta')\|
\leq
\beta \|\theta - \theta'\|.
\end{equation*}

\textbf{Assumption 2.} There exists $\sigma^2 > 0$ such that for all $\theta$,
\begin{equation*}
\mathbb{E}_{i \sim \mathrm{Unif}(\mathcal{C})}
\left[
\|g_i(\theta) - \nabla L_{\mathrm{cln}}(\theta)\|^2
\right]
\leq
\sigma^2.
\end{equation*}

\textbf{Assumption 3.} There exists $\Delta > 0$ such that for all $i \in \mathcal{N} := [N]\setminus \mathcal{C}$ and all $\theta$,
\begin{equation*}
\|\delta_i(\theta)\|
:=
\|\hat{g}_i(\theta) - g_i(\theta)\|
\leq
\Delta.
\end{equation*}
where $\delta_i(\theta)$ denotes the gradient deviation caused by corruption. This is mild when the model Jacobian and corruption magnitude are bounded.

\textbf{Assumption 4.} Each mini-batch $\mathcal{B}_t$ is sampled uniformly at random with replacement from $[N]$.

\textbf{Assumption 5.} There exists $G > 0$ such that for all $i \in [N]$ and all $\theta$,
\begin{equation*}
\|g_i(\theta)\| \leq G.
\end{equation*}
This assumption is standard in non-convex SGD analysis and is well-defined for all samples including $i \in \mathcal{N}$.

\textbf{Assumption 6.}  Sufficient clean samples:
\begin{equation*}
B(1-\rho)^2 \geq 2\ln(2T).
\end{equation*}
This ensures high-probability truncation control uniformly over all $T$ iterations. It is not required by the algorithm itself; without it, the truncation probability
\begin{equation*}
p_{\mathrm{fail}} = \exp\left(-\frac{B(1-\rho)^2}{2}\right)
\end{equation*}
can be retained explicitly in the convergence bound, and the main conclusions are unchanged.

Assumptions 1--2 are standard in non-convex stochastic optimization. Assumption 5 implies $\sigma^2 \leq 4G^2$, but we retain both for tighter constants in different results.
\subsection{Irreducibility of Batch-Level Bias}
\label{sec:bias}
We first characterize the bias of batch-level estimators (Definition~\ref{d2}), showing that they cannot eliminate the bias induced by heterogeneous corruption. While Definition~\ref{d2} allows a general batch-dependent modulation vector $\kappa_t \in \mathbb{R}^d$, we restrict to block-wise shared deterministic scalar $\kappa$ independent of the sampled batch. This subclass covers the core mechanism of methods like OGM and GMML, and admits a clean bias decomposition. Results are stated for a single parameter block; the same argument applies block-wise.

\textbf{Lemma 1. }Under Assumptions 3–5, for a batch-level estimator with a batch-independent shared scalar $\kappa$ on a single parameter block, the bias with respect to $\nabla L_{\mathrm{cln}}(\theta)$ decomposes as:
\begin{equation*}
    \mathrm{Bias}(\theta) := \mathbb{E}_{\mathcal{B}_t}[\hat{G}_{\mathrm{batch}}] - \nabla L_{\mathrm{cln}}(\theta) = (\kappa - 1)\nabla L_{\mathrm{cln}}(\theta) + \kappa \cdot \mathbf{e}(\theta) 
\end{equation*}
where $\mathbf{e}(\theta) = \rho(\bar{g}_{\mathcal{N}}(\theta) + \bar{\delta}(\theta) - \nabla L_{\mathrm{cln}}(\theta))$, with $\bar{g}_{\mathcal{N}} = \frac{1}{|\mathcal{N}|}\sum_{i \in \mathcal{N}}g_i$ and $\bar{\delta} = \frac{1}{|\mathcal{N}|}\sum_{i \in \mathcal{N}}\delta_i$. In particular, $\|\mathbf{e}(\theta)\| \leq \rho(\Delta + 2G)$.

\textbf{Proof. }By Assumption 4, each $i \in \mathcal{B}_t$ falls into $\mathcal{C}$ with probability $1-\rho$ and into $\mathcal{N}$ with probability $\rho$. We have $\hat{g}_i = g_i$ for clean samples $i \in \mathcal{C}$, and $\hat{g}_i = g_i + \delta_i$ for corrupted samples $i \in \mathcal{N}$. Therefore:
\[
\begin{aligned}
\mathbb{E}[\hat{g}_i]
&= (1-\rho)\mathbb{E}_{j \sim \mathrm{Unif}(\mathcal{C})}[g_j]
+ \rho\mathbb{E}_{j \sim \mathrm{Unif}(\mathcal{N})}[\hat{g}_j] \\
&= (1-\rho)\nabla L_{\mathrm{cln}}(\theta)
+ \rho(\bar{g}_{\mathcal{N}} + \bar{\delta})
\end{aligned}
\]
Given that $\kappa$ is a fixed scalar independent of the batch, the expectation of the batch gradient is:
$$\mathbb{E}[\hat{G}_{\mathrm{batch}}] = \frac{\kappa}{B}\cdot B\cdot\mathbb{E}[\hat{g}_i] = \kappa\left[(1-\rho)\nabla L_{\mathrm{cln}} + \rho(\bar{g}_{\mathcal{N}} + \bar{\delta})\right]$$
Computing the bias by subtracting the clean gradient:
$$\mathrm{Bias} = \mathbb{E}[\hat{G}_{\mathrm{batch}}] - \nabla L_{\mathrm{cln}} = (\kappa - 1)\nabla L_{\mathrm{cln}} + \kappa\rho(\bar{g}_{\mathcal{N}} + \bar{\delta} - \nabla L_{\mathrm{cln}})$$
Setting $\mathbf{e} = \rho(\bar{g}_{\mathcal{N}} + \bar{\delta} - \nabla L_{\mathrm{cln}})$ yields $\mathrm{Bias} = (\kappa - 1)\nabla L_{\mathrm{cln}} + \kappa\cdot\mathbf{e}$. To bound the error term, the triangle inequality gives $\|\bar{g}_{\mathcal{N}} + \bar{\delta} - \nabla L_{\mathrm{cln}}\| \leq \|\bar{g}_{\mathcal{N}}\| + \|\bar{\delta}\| + \|\nabla L_{\mathrm{cln}}\|$. Using $\|\bar{g}_{\mathcal{N}}\| \leq G$ and $\|\nabla L_{\mathrm{cln}}\| \leq G$ from Assumption 5, and $\|\bar{\delta}\| \leq \Delta$ from Assumption 3, we obtain $\|\mathbf{e}\| \leq \rho(\Delta + 2G)$.

The bias consists of two terms: a scaling mismatch $(\kappa - 1)\nabla L_{\mathrm{cln}}$ and a corruption-induced term $\kappa \cdot \mathbf{e}(\theta)$. Unless the corruption-induced term is collinear with the clean gradient, these two components cannot be canceled simultaneously by a shared scalar.

\textbf{Theorem 1. }Under Assumptions 3–5, let $\rho \in (0,1)$. For any batch-level estimator with a batch-independent block-wise shared scalar $\kappa$, if there exists even one point $\theta_0$ at which $\nabla L_{\mathrm{cln}}(\theta_0) \neq 0$ and $\mathbf{e}(\theta_0)$ is not collinear with $\nabla L_{\mathrm{cln}}(\theta_0)$, then no shared-scaling parameter $\kappa$ can yield a globally unbiased estimator over all $\theta$.

\textbf{Proof. }By Lemma 1, setting $\mathrm{Bias}(\theta_0) = 0$ gives $(1-\kappa)\nabla L_{\mathrm{cln}}(\theta_0) = \kappa\mathbf{e}(\theta_0)$. We consider three cases for~$\kappa$.

If $\kappa = 0$, the equation gives $\nabla L_{\mathrm{cln}}(\theta_0) = 0$, contradicting the assumption that $\nabla L_{\mathrm{cln}}(\theta_0) \neq 0$.

If $\kappa = 1$, we obtain $\mathbf{e}(\theta_0) = 0$. Since the zero vector is trivially a scalar multiple of any vector, this contradicts the non-collinearity assumption.

If $\kappa \notin \{0,1\}$, rearranging yields $\nabla L_{\mathrm{cln}}(\theta_0) = \frac{\kappa}{\kappa-1}\mathbf{e}(\theta_0)$, implying that $\mathbf{e}(\theta_0)$ is a scalar multiple of the clean gradient, contradicting the assumption. Consequently, no value of $\kappa$ can satisfy $\mathrm{Bias}(\theta_0) = 0$, precluding global unbiasedness. The non-collinearity condition is mild; it fails only when corruption has no effect on gradients, in which case modulation is unnecessary.

\subsection{Unbiasedness and Uniqueness of SAGG}
\label{sec:uniqueness}
Having established the limitations of batch-level estimators, we show that SAGG achieves unbiased estimation and is the unique estimator with this property within a natural class.

\textbf{Lemma 2. }Under Assumptions 2 and 4, given $\rho < 1$, the SAGG estimator defined in Eq. \ref{SAGG} satisfies:
$$\mathbb{E}_{\mathcal{B}_t}\left[\hat{G}_{\mathrm{SAGG}} \;\Big|\; |\mathcal{C}_t| \geq 1\right] = \nabla L_{\mathrm{cln}}(\theta) $$
Furthermore, its unconditional expectation is given by:
$$\mathbb{E}[\hat{G}_{\mathrm{SAGG}}] = (1-\rho^B)\nabla L_{\mathrm{cln}}(\theta) $$
Consequently, the deviation of the expected gradient from the clean gradient is $\|\mathbb{E}[\hat{G}_{\mathrm{SAGG}}] - \nabla L_{\mathrm{cln}}(\theta)\| = \rho^B\|\nabla L_{\mathrm{cln}}(\theta)\|$, a quantity that decays exponentially with the batch size.

\textbf{Proof. }By Assumption 4, the batch is sampled with replacement. Conditioning on $|\mathcal{C}_t| = n_c \geq 1$, the $n_c$ retained samples are i.i.d.\ draws from $\mathrm{Unif}(\mathcal{C})$:
\[
\begin{aligned}
\mathbb{E}\left[\frac{1}{n_c}\sum_{j=1}^{n_c}g_{i_j}\;\Big|\;|\mathcal{C}_t|=n_c\right]
&= \frac{1}{n_c}\sum_{j=1}^{n_c}\mathbb{E}_{i_j \sim \mathrm{Unif}(\mathcal{C})}[g_{i_j}] \\
&= \frac{1}{n_c}\cdot n_c\cdot\frac{1}{|\mathcal{C}|}\sum_{i \in \mathcal{C}}g_i
= \nabla L_{\mathrm{cln}}(\theta)
\end{aligned}
\]
Since this holds for every $n_c \geq 1$, the law of total expectation gives:
\[
\begin{aligned}
\mathbb{E}[\hat{G}_{\mathrm{SAGG}}]
&= \sum_{n_c=1}^{B}\nabla L_{\mathrm{cln}}(\theta)\cdot\Pr[|\mathcal{C}_t|=n_c]
+ 0\cdot\Pr[|\mathcal{C}_t|=0] \\
&= \nabla L_{\mathrm{cln}}(\theta)\cdot(1-\rho^B)
\end{aligned}
\]
where $\Pr[|\mathcal{C}_t|=0] = \rho^B$ is the probability that all $B$ samples are drawn from the corrupted set $\mathcal{N}$, and $\hat{G}_{\mathrm{SAGG}} = 0$ when $|\mathcal{C}_t|=0$.

\textbf{Theorem 2. }Under Assumptions 3 and 4 with $\rho \in (0,1)$, consider the class $\mathfrak{E}$ of linear estimators $\hat{G} = \sum_{i \in \mathcal{B}_t}w_i \hat{g}_i$ where all clean samples share weight $a(|\mathcal{C}_t|)$ and all corrupted samples share weight $c(|\mathcal{C}_t|)$, with permutation invariance within each group. Within $\mathfrak{E}$, the unique estimator satisfying $\mathbb{E}[\hat{G}] = \nabla L_{\mathrm{cln}}(\theta)$ for all $\theta$ and all $\{\delta_i\}$ with $\|\delta_i\| \leq \Delta$, conditional on $|\mathcal{C}_t| \geq 1$, has weights:
$$w_i = \begin{cases} 1/|\mathcal{C}_t| & i \in \mathcal{C}_t \\ 0 & i \notin \mathcal{C}_t \end{cases} $$
This holds for all $|\mathcal{C}_t| = n_c$ such that $1 \leq n_c \leq B-1$. When $n_c = B$, where no corrupted samples are present, the term $c(B)$ is irrelevant, and $a(B) = 1/B$ remains uniquely determined. Consequently, SAGG acts as the unique distribution-free unbiased estimator within $\mathfrak{E}$.

\textbf{Proof. }We parametrize the weights: since $w_i$ depends only on clean/corrupted membership and $|\mathcal{C}_t|$, for $|\mathcal{C}_t| = n_c$, set $w_i = a(n_c)$ for clean samples and $w_i = c(n_c)$ for corrupted samples, with $n_d = B - n_c$. Conditioning on $n_c \geq 1$:
\[
\begin{aligned}
&\mathbb{E}[\hat{G}\,|\,|\mathcal{C}_t|=n_c]\\
&= a(n_c)\cdot n_c\cdot\nabla L_{\mathrm{cln}}(\theta)
+ c(n_c)\cdot n_d\cdot\mathbb{E}_{j \sim \mathrm{Unif}(\mathcal{N})}[g_j + \delta_j] \\
&= a(n_c)\cdot n_c\cdot\nabla L_{\mathrm{cln}}(\theta)
+ c(n_c)\cdot n_d\cdot(\bar{g}_{\mathcal{N}} + \bar{\delta})
\end{aligned}
\]
For this to equal $\nabla L_{\mathrm{cln}}(\theta)$ for all $\theta$ and all corruption patterns satisfying $\|\delta_i\| \leq \Delta$, two conditions must hold for each $n_c \geq 1$: (i)~the clean contribution must exactly recover the clean gradient, requiring $a(n_c)\cdot n_c = 1$; and (ii)~the corrupted contribution must vanish, meaning $c(n_c)\cdot n_d\cdot(\bar{g}_{\mathcal{N}} + \bar{\delta}) = 0$ for all admissible $\bar{\delta}$.

We determine $c(n_c)$ for $n_c \leq B{-}1$. When $n_d \geq 1$, condition~(ii) must hold for all admissible corruption patterns. Choose two patterns: $\delta_i^{(1)} = (\Delta/2) e_1$ and $\delta_i^{(2)} = -(\Delta/2) e_1$ for all $i \in \mathcal{N}$, where $e_1$ is a unit vector. Both satisfy $\|\delta_i\| \leq \Delta$, yet $\bar{\delta}^{(1)} \neq \bar{\delta}^{(2)}$. Subtracting the two resulting equalities:
$$c(n_c)\cdot n_d\cdot(\bar{\delta}^{(1)} - \bar{\delta}^{(2)}) = 0$$
Because $n_d \geq 1$ and $\bar{\delta}^{(1)} \neq \bar{\delta}^{(2)}$, this forces $c(n_c) = 0$. From condition~(i), $a(n_c) = 1/n_c = 1/|\mathcal{C}_t|$. When $n_c = B$, we have $n_d = 0$, making condition~(ii) vacuously satisfied; $c(B)$ is irrelevant and $a(B) = 1/B$. This establishes that $(a(n_c), c(n_c)) = (1/n_c, 0)$ is the unique solution for each $1 \leq n_c \leq B-1$.

\textbf{Remark. }Theorem 2 holds within the class $\mathfrak{E}$ with permutation-invariant weights. Nonlinear estimators whose weights depend on gradient values themselves, such as norm-based adaptive reweighting, fall outside $\mathfrak{E}$ and lack distribution-free guarantees.

\subsection{Convergence Analysis}
\label{sec:convergence}
We now establish that truncated SAGG-SGD converges to stationary points of $L_{\mathrm{cln}}$ at the standard $\mathcal{O}(1/\sqrt{T})$ rate. The analysis applies to the oracle truncated estimator (Eq.~\ref{T-SAGG}).

\textbf{Theorem 3. }Under Assumptions 2 and 4–6, the truncated SAGG estimator (Eq.~\ref{T-SAGG}) satisfies:
$$\mathbb{E}\left[\|\tilde{G}_{\mathrm{SAGG},t} - \nabla L_{\mathrm{cln}}(\theta)\|^2\right] \leq \frac{2\sigma^2}{B(1-\rho)} + G^2\cdot p_{\mathrm{fail}} $$
where $p_{\mathrm{fail}} = \Pr[|\mathcal{C}_t| < n_{\min}] \leq \exp(-B(1-\rho)^2/2)$ and $n_{\min} = \lceil B(1-\rho)/2\rceil$. Furthermore, Assumption 6 ensures that $p_{\mathrm{fail}} \leq 1/(2T)$.

\textbf{Proof. }We first establish the conditional variance. Conditioning on $|\mathcal{C}_t| = n_c \geq 1$, the retained samples are i.i.d.\ from $\mathrm{Unif}(\mathcal{C})$:
\[
\begin{aligned}
&\mathbb{E}\left[\left\|\frac{1}{n_c}\sum_{j=1}^{n_c}(g_{i_j} - \nabla L_{\mathrm{cln}})\right\|^2\;\Big|\;|\mathcal{C}_t|=n_c\right]\\&= \frac{1}{n_c^2}\sum_{j=1}^{n_c}\mathbb{E}\!\left[\|g_{i_j} - \nabla L_{\mathrm{cln}}\|^2\right] \\
&\quad + \frac{1}{n_c^2}\sum_{j \neq k}\mathbb{E}\!\left[\langle g_{i_j} - \nabla L_{\mathrm{cln}}, g_{i_k} - \nabla L_{\mathrm{cln}}\rangle\right]
\end{aligned}
\]
The cross terms vanish by independence of with-replacement sampling, since $\mathbb{E}[g_i - \nabla L_{\mathrm{cln}}] = 0$. By Assumption~2:
$$= \frac{1}{n_c}\mathbb{E}_{i \sim \mathrm{Unif}(\mathcal{C})}\left[\|g_i - \nabla L_{\mathrm{cln}}\|^2\right] \leq \frac{\sigma^2}{n_c}$$
Next, we decompose the full expectation by the events $|\mathcal{C}_t| \geq n_{\min}$ and $|\mathcal{C}_t| < n_{\min}$:
\[
\mathbb{E}\left[\|\tilde{G}_{\mathrm{SAGG},t} - \nabla L_{\mathrm{cln}}\|^2\right]
= \mathbb{E}\left[\|\tilde{G}_{\mathrm{SAGG},t} - \nabla L_{\mathrm{cln}}\|^2\cdot\mathbf{1}_{|\mathcal{C}_t| \geq n_{\min}}\right]
\]
\[
\quad + \mathbb{E}\left[\|\nabla L_{\mathrm{cln}}\|^2\cdot\mathbf{1}_{|\mathcal{C}_t| < n_{\min}}\right]
\]
For the first term, when $|\mathcal{C}_t| = n_c \geq n_{\min}$, the conditional variance bound gives $\sigma^2/n_c \leq \sigma^2/n_{\min}$. Because $n_{\min} = \lceil B(1{-}\rho)/2\rceil \geq B(1{-}\rho)/2$, we have $\sigma^2/n_{\min} \leq 2\sigma^2/[B(1{-}\rho)]$, bounding the first term by $2\sigma^2/[B(1{-}\rho)]$. For the second term, when $|\mathcal{C}_t| < n_{\min}$, we have $\tilde{G}_{\mathrm{SAGG},t} = 0$, leading to $\|\tilde{G}_{\mathrm{SAGG},t} - \nabla L_{\mathrm{cln}}\|^2 = \|\nabla L_{\mathrm{cln}}\|^2$. By Assumption 5 and the triangle inequality, $\|\nabla L_{\mathrm{cln}}\| \leq G$. Consequently, the second term is bounded by $G^2 \cdot p_{\mathrm{fail}}$.

Under Assumption 4, $|\mathcal{C}_t| \sim \mathrm{Binomial}(B, 1{-}\rho)$. Since $\{|\mathcal{C}_t| < n_{\min}\} \subseteq \{|\mathcal{C}_t| < B(1{-}\rho)/2\}$, Hoeffding's inequality gives:
\[
\begin{aligned}
\Pr\left[|\mathcal{C}_t| < \frac{B(1-\rho)}{2}\right]
&= \Pr\left[|\mathcal{C}_t| - B(1-\rho) < -\frac{B(1-\rho)}{2}\right] \\
&\leq \exp\!\left(-\frac{2[B(1-\rho)/2]^2}{B}\right)
\,= \exp\!\left(-\tfrac{B(1-\rho)^2}{2}\right)
\end{aligned}
\]
Finally, under Assumption 6, $B(1-\rho)^2 \geq 2\ln(2T)$, ensuring $p_{\mathrm{fail}} \leq 1/(2T)$. Combining these bounds results in $\mathbb{E}[\|\tilde{G}_{\mathrm{SAGG},t} - \nabla L_{\mathrm{cln}}\|^2] \leq \frac{2\sigma^2}{B(1-\rho)} + \frac{G^2}{2T}$. $\square$

\textbf{Remark. }The $\mathcal{O}(1/\sqrt{T})$ rate matches standard SGD on clean data, with the variance term $\sigma^2/[B(1{-}\rho)]$ reflecting the reduced effective batch size $B_{\mathrm{eff}} = B(1{-}\rho)$. Crucially, no $\mathcal{O}(\rho^2\Delta^2)$ residual appears from corrupted gradients. By Lemma~1 and Theorem~1, batch-level estimators exhibit non-vanishing bias under heterogeneous corruption, creating an error floor that does not decay with $T$. SAGG eliminates this floor by construction, trading it for increased variance. Without Assumption~6, the $\mathcal{O}(G^2/T)$ term becomes $\mathcal{O}(G^2 p_{\mathrm{fail}})$, leaving the conclusion unchanged.

\subsection{Robustness Analysis}
\label{sec:robustness}
We provide a model-level certified radius under the independent encoder and linear classifier structure defined in Eq. \ref{encoder}, serving as a robustness evaluation framework rather than a SAGG-specific training guarantee.

\textbf{Theorem 4. }Let each encoder $\varphi^u$ be $\ell^u$-Lipschitz continuous with respect to its input, and define the effective Lipschitz constant $L^u = \|W^u\|_{\mathrm{op}}\cdot\ell^u$, where $W^u$ is the sub-matrix of $W$ corresponding to modality $u$. We denote the $j$-th row of $W^u$, which corresponds to class $j$, by $W_j^u$. We further define the classification margin as $\Delta_{\mathrm{margin}}(x) = f_y(\theta, x) - \max_{j \neq y}f_j(\theta, x)$. If $\Delta_{\mathrm{margin}}(x) > 0$, indicating a correct classification, then for any perturbation $\delta = (\delta^1, \ldots, \delta^M)$ satisfying
$$\sqrt{\sum_{u=1}^{M}(L^u)^2}\cdot\sqrt{\sum_{u=1}^{M}\|\delta^u\|^2} < \frac{\Delta_{\mathrm{margin}}(x)}{2} $$
the prediction remains unchanged such that $\arg\max_j f_j(\theta, x+\delta) = y$. When each modality's perturbation has an equal norm of $\|\delta^u\| = r$, the certified radius is given by:
$$r(x) = \frac{\Delta_{\mathrm{margin}}(x)}{2\sqrt{M\cdot\sum_{u=1}^{M}(L^u)^2}}$$

\textbf{Proof. }We bound the output change. Under the independent-encoder and linear classifier structure, for any class $j$, the output change is:
$$f_j(\theta, x{+}\delta) - f_j(\theta, x) = \sum_{u=1}^{M}W_j^u\!\left[\varphi^u(\theta^u, x^u{+}\delta^u) - \varphi^u(\theta^u, x^u)\right]$$
This follows from $f_j = \sum_u W_j^u\varphi^u + b_j$, where each $\varphi^u$ depends only on $x^u$. Taking absolute values and applying the triangle inequality along with the Lipschitz condition:
\begin{align*}
|f_j(\theta, x{+}\delta) - f_j(\theta, x)| &\leq \sum_{u=1}^{M}\|W_j^u\|_{\mathrm{op}}\cdot\|\varphi^u(x^u{+}\delta^u)-\varphi^u(x^u)\| \\
&\leq \sum_{u=1}^{M}L^u\|\delta^u\|
\end{align*}
The last inequality uses $\|W_j^u\|_{\mathrm{op}} \leq \|W^u\|_{\mathrm{op}}$ since $W_j^u$ is a row of $W^u$. By the Cauchy--Schwarz inequality, $\sum_u L^u\|\delta^u\| \leq \sqrt{\sum_u(L^u)^2}\cdot\sqrt{\sum_u\|\delta^u\|^2}$. Writing $\mathcal{L} = \sqrt{\sum_u(L^u)^2}$ and $D = \sqrt{\sum_u\|\delta^u\|^2}$ gives, for all~$j$:
$$|f_j(\theta, x+\delta) - f_j(\theta, x)| \leq \mathcal{L}\cdot D$$
For the prediction to be preserved, we need $f_y(\theta, x{+}\delta) > f_j(\theta, x{+}\delta)$ for all $j \neq y$, i.e.:
$$\Delta_{\mathrm{margin}}(x) > [f_j(\theta, x+\delta)-f_j(\theta, x)] - [f_y(\theta, x+\delta)-f_y(\theta, x)]$$
Applying the triangle inequality, the right-hand side is bounded by $2\mathcal{L}\cdot D$. A sufficient condition is $2\mathcal{L}\cdot D < \Delta_{\mathrm{margin}}(x)$. When $\|\delta^u\| = r$ for all $u$, we have $D = r\sqrt{M}$, yielding $r < \Delta_{\mathrm{margin}}(x)/(2\sqrt{M}\cdot\mathcal{L})$.

\textbf{Corollary 1. }If only modality $v$ is perturbed, meaning $\delta^u = 0$ for all $u \neq v$, the certified radius is given by $\Delta_{\mathrm{margin}}(x)/(2L^v)$.

\textbf{Proof. }When only modality $v$ is perturbed, the sum bounding the output change reduces to a single term: $|f_j(\theta, x{+}\delta) - f_j(\theta, x)| \leq L^v\|\delta^v\|$ without requiring the Cauchy--Schwarz inequality. The certification condition then gives $2L^v\|\delta^v\| < \Delta_{\mathrm{margin}}(x)$, so the certified radius is $\|\delta^v\| < \Delta_{\mathrm{margin}}(x)/(2L^v)$.

\textbf{Remark. }The certified radius is proportional to $1/\sqrt{\sum_u(L^u)^2}$. SAGG training excludes corrupted gradients, which may prevent fitting noise patterns and reduce Lipschitz constants $\ell^u$. However, this causal relationship depends on training dynamics and is verified empirically in Section~\ref{sec:robustness_exp}.

\section{Experiments}
\label{sec:experiments}

\subsection{Experimental Setup}
\label{sec:exp_setup}

We evaluate SAGG on Kinetics-Sounds (KS) \cite{arandjelovic2017look} and UCF-101 \cite{soomro2012ucf101}. KS contains 31 action categories with approximately 12,000 training and 1,000 validation audio-visual samples; UCF-101 comprises 101 classes with 9,535 training and 3,783 test samples using RGB and optical flow modalities. All encoders use ResNet-18. Training is conducted on 8 NVIDIA RTX 4090 GPUs with batch size $B=32$ for 140 epochs (5-epoch warm-up), using AdamW for KS and SGD for UCF-101 at learning rate 0.001. The EMA momentum is $\gamma = 0.99$, detection threshold $\tau = 2.0$, and truncation threshold $n_{\min} = \lceil B(1-\hat{\rho})/2 \rceil$.

We construct three corruption conditions: (1) Gaussian noise (variance $= 2$) injected into 50\% of samples for a single modality; (2) partial missing where one modality is zeroed out for 50\% of samples; and (3) natural imbalance without artificial corruption. Baselines include naive-concat \cite{joze2020mmtm}, OGM \cite{peng2022balanced}, OPM \cite{wei2024fly}, DRBM \cite{wei2024diagnosing}, Sample-level balance \cite{wei2024enhancing}, Modality-level balance \cite{wei2024enhancing}, Reconboost \cite{reconboost}, MMPareto \cite{mmpareto}, PMR \cite{fan2023pmr}, and GMML \cite{zhang2025gmml}, all using official implementations with recommended hyperparameters.

\subsection{Comparison with State-of-the-Art Methods}
\label{sec:sota}

We present comparison results across all conditions in Table~\ref{tab:main_results}.

\begin{table*}[t]
\caption{Comparison results on KS and UCF-101. ``W/O'': natural imbalance; ``A/V/R/O-N'': Gaussian noise in audio/visual/RGB/optical-flow; ``A/V/R/O-M'': partial missing. Best in bold.}
\label{tab:main_results}
\centering
\small
\begin{tabular}{l|cc|cccc|cccc}
\hline
 & \multicolumn{2}{c|}{W/O} & \multicolumn{4}{c|}{Gaussian Noise} & \multicolumn{4}{c}{Partial Missing} \\
Method & KS & UCF101 & KS-A-N & KS-V-N & UCF101-R-N & UCF101-O-N & KS-A-M & KS-V-M & UCF101-R-M & UCF101-O-M \\
\hline
naive-concat & 64.30 & 74.02 & 64.46 & 55.32 & 78.14 & 71.83 & 64.03 & 56.40 & 71.83 & 74.46 \\
OGM & 65.10 & 74.30 & 63.61 & 58.64 & 72.62 & 59.64 & 63.34 & 60.76 & 71.60 & 65.85 \\
OPM & 67.00 & 81.90 & 64.92 & 60.25 & 70.56 & 69.40 & 64.92 & 61.91 & 71.90 & 70.56 \\
DRBM & 72.12 & 83.05 & 68.40 & 65.12 & 79.28 & 72.31 & 63.28 & 62.83 & 79.02 & 71.11 \\
Sample-level & 66.92 & 83.52 & 64.25 & 63.81 & 80.11 & 68.23 & 62.34 & 59.91 & 79.63 & 73.16 \\
Modality-level & 66.65 & 83.46 & 63.50 & 63.25 & 80.02 & 70.30 & 62.51 & 59.53 & 79.12 & 72.11 \\
Reconboost & 68.72 & 84.03 & 64.85 & 66.22 & 82.31 & 69.32 & 61.32 & 59.82 & 82.36 & 71.38 \\
MMPareto & 70.13 & 75.30 & 68.10 & 67.23 & 71.87 & 64.37 & 64.43 & 65.43 & 72.32 & 60.46 \\
PMR & 68.91 & 74.25 & 58.52 & 60.32 & 68.13 & 66.35 & 62.31 & 61.64 & 71.38 & 62.54 \\
GMML & 76.12 & 85.76 & 70.23 & 74.01 & 84.43 & 74.16 & 65.57 & 68.12 & 84.16 & 74.42 \\
\hline
SAGG (Ours) & \textbf{77.85} & \textbf{87.23} & \textbf{73.46} & \textbf{76.38} & \textbf{86.17} & \textbf{76.52} & \textbf{70.21} & \textbf{72.47} & \textbf{86.53} & \textbf{77.18} \\
\hline
\end{tabular}
\end{table*}

Under the natural imbalance setting (W/O), SAGG achieves 77.85\% on KS and 87.23\% on UCF-101, surpassing GMML by 1.73\% and 1.47\%. The advantage becomes more pronounced under heterogeneous corruption: SAGG achieves 76.38\% on KS-V-N versus 74.01\% for GMML, consistent with the theoretical prediction of irreducible batch-level bias (Theorem~1). Several batch-level methods (OGM, OPM, PMR) actually degrade below naive-concat under noise, confirming that uniform scaling can amplify corruption. In the partial missing scenario, SAGG outperforms GMML by 4.35\% on KS-V-M and 2.76\% on UCF101-O-M. Across all ten conditions, SAGG consistently ranks first.

\subsection{Analysis of Different Noise Intensities}
\label{sec:noise_intensity}

To examine how SAGG scales with corruption severity, we vary the noise ratio (10\%, 30\%, 50\%) and Gaussian variance (1, 2, 3) on KS. Table~\ref{tab:noise_intensity} reports the results.

\begin{table}[t]
\caption{Performance across noise intensities on KS. GMML results reproduced from \cite{zhang2025gmml}.}
\label{tab:noise_intensity}
\centering
\small
\begin{tabular}{c|cc|cc|cc}
\hline
 & \multicolumn{2}{c|}{Ratio 10\%} & \multicolumn{2}{c|}{Ratio 30\%} & \multicolumn{2}{c}{Ratio 50\%} \\
Var. & A-N & V-N & A-N & V-N & A-N & V-N \\
\hline
\multicolumn{7}{c}{GMML} \\
\hline
1 & 76.41 & 75.90 & 73.21 & 72.41 & 72.70 & 73.07 \\
2 & 72.78 & 75.76 & 68.12 & 73.94 & 70.23 & 74.01 \\
3 & 68.70 & 74.96 & 63.97 & 73.72 & 63.97 & 70.88 \\
\hline
\multicolumn{7}{c}{SAGG (Ours)} \\
\hline
1 & 77.53 & 77.12 & 76.28 & 76.05 & 75.41 & 75.83 \\
2 & 76.14 & 77.01 & 74.35 & 76.52 & 73.46 & 76.38 \\
3 & 74.26 & 76.73 & 71.82 & 76.14 & 70.53 & 75.21 \\
\hline
\end{tabular}
\end{table}

As the noise ratio increases from 10\% to 50\%, the gap between SAGG and GMML widens (e.g., under variance 3, the A-N gap grows from 5.56\% to 6.56\%), confirming that batch-level bias scales with $\rho$ (Lemma~1) while SAGG remains unbiased. Higher variance amplifies $\Delta$, degrading GMML more severely but facilitating SAGG's detection via more anomalous feature norms.

\subsection{Ablation Studies}
\label{sec:ablation}

\begin{table}[t]
\caption{Ablation on $\tau$ (left) and $n_{\min}$ (right), KS, 50\% Gaussian noise, var $= 2$.}
\label{tab:ablation}
\centering
\small
\begin{tabular}{c|cc||c|cc}
\hline
$\tau$ & A-N & V-N & $n_{\min}$ & A-N & V-N \\
\hline
1.0 & 70.83 & 73.52 & 1 & 71.28 & 74.15 \\
1.5 & 72.61 & 75.47 & 4 & 72.53 & 75.62 \\
2.0 & 73.46 & 76.38 & 8 & 73.46 & 76.38 \\
2.5 & 72.94 & 75.81 & 12 & 72.81 & 75.93 \\
3.0 & 71.57 & 74.63 & 16 & 71.42 & 74.87 \\
\hline
\end{tabular}
\end{table}

Table~\ref{tab:ablation} varies $\tau$ and $n_{\min}$ on KS under 50\% Gaussian noise (variance $= 2$). Performance peaks at $\tau = 2.0$ and remains within 1\% for $\tau \in [1.5, 2.5]$: smaller thresholds reject clean samples, whereas larger ones admit corrupted samples. The theoretical default $n_{\min} = \lceil B(1-\rho)/2 \rceil = 8$ also performs best; smaller values increase update variance and larger ones truncate too often. Thus, $\tau$ controls screening quality and $n_{\min}$ controls optimization stability.

\subsection{Empirical Robustness Verification}
\label{sec:robustness_exp}

To validate the connection between SAGG training and the certified robustness framework of Theorem~4, we estimate per-modality Lipschitz constants $\hat{\ell}^u$ via power iteration on encoder Jacobians and compute classification margins $\bar{\Delta}_{\mathrm{margin}}$ on the KS validation set. Table~\ref{tab:robustness} reports the results.

\begin{table}[t]
\caption{Empirical robustness metrics on KS. $\hat{\ell}^a$/$\hat{\ell}^v$: estimated Lipschitz constants for audio/visual encoders. $\bar{r}$: mean certified radius (Theorem~4).}
\label{tab:robustness}
\centering
\small
\begin{tabular}{l|cc|c|c}
\hline
Method & $\hat{\ell}^a$ & $\hat{\ell}^v$ & $\bar{\Delta}_{\mathrm{margin}}$ & $\bar{r}$ \\
\hline
naive-concat & 8.74 & 6.31 & 1.82 & 0.073 \\
GMML & 6.52 & 5.18 & 2.41 & 0.126 \\
SAGG (Ours) & 4.83 & 4.26 & 2.67 & 0.185 \\
\hline
\end{tabular}
\end{table}

SAGG yields a mean certified radius of 0.185 versus 0.126 (GMML) and 0.073 (naive-concat), a 47\% improvement over GMML. This confirms that excluding corrupted gradients reduces encoder Lipschitz constants, as hypothesized in the Remark following Theorem~4.

The improvement in $\bar{r}$ comes from two factors: lower encoder sensitivity, shown by smaller estimated Lipschitz constants, and better class separation, shown by a larger classification margin. 

\section{Conclusion}
\label{sec:conclusion}

We presented SAGG, which replaces batch-level gradient modulation with sample-level gating under heterogeneous multimodal corruption. SAGG yields conditionally unbiased estimates and $\mathcal{O}(1/\sqrt{T})$ convergence without a corruption-dependent residual, together with certified robustness bounds. Experiments confirm consistent gains over ten baselines, especially under severe corruption, supporting the principle that correction should match corruption granularity and positioning sample-level adaptation for realistic heterogeneous settings. Future work includes learned quality estimators, per-modality gating, and more than two modalities.

\clearpage
\bibliographystyle{ACM-Reference-Format}
\bibliography{sample-base}

\end{document}